\documentclass{article}
\usepackage{spconf,amsmath,graphicx}

\usepackage{xcolor}
\usepackage{caption}
\usepackage{booktabs}
\usepackage{multirow} 

\usepackage{algorithm}
\usepackage{algorithmic}
\usepackage{graphicx}

\usepackage{appendix}

\title{Bag of Tricks with Quantized Convolutional Neural Networks for image classification}
\name{
Jie Hu\textsuperscript{\rm {1,2*}} \thanks{$^*$Equal contribution.},
Mengze Zeng\textsuperscript{\rm {3*}},
Enhua Wu\textsuperscript{\rm {1,2,4$\dagger$}} \thanks{$^\dagger$Corresponding author. This work is supported in part by NSFC Grants (62072449).}
}
\address{$^1$State Key Laboratory of Computer Science, ISCAS\\
$^2$University of Chinese Academy of Sciences \\
$^3$ByteDance Inc. \hspace{9mm}
$^4$University of Macau \\
\normalsize{hujie@ios.ac.cn, \; zeitzmz@gmail.com, \; ehwu@um.edu.mo}
}

\begin{document}
%
\maketitle

\begin{abstract}
	
	Deep neural networks have been proven effective in a wide range of tasks. However, their high computational and memory costs make them impractical to deploy on resource-constrained devices. To address this issue, quantization schemes have been proposed to reduce the memory footprint and improve inference speed. While numerous quantization methods have been proposed, they lack systematic analysis for their effectiveness. 
    To bridge this gap, we collect and improve existing quantization methods and propose a gold guideline for post-training quantization. We evaluate the effectiveness of our proposed method with two popular models, ResNet50 and MobileNetV2, on the ImageNet dataset. By following our guidelines, no accuracy degradation occurs even after directly quantizing the model to 8-bits without additional training. 
    A quantization-aware training based on the guidelines can further improve the accuracy in lower-bits quantization. Moreover, we have integrated a multi-stage fine-tuning strategy that works harmoniously with existing pruning techniques to reduce cost even further. Remarkably, our results reveal that a quantized MobileNetV2 with 30\% sparsity actually surpasses the performance of the equivalent full-precision model, underscoring the effectiveness and resilience of our proposed scheme.

\end{abstract}
\vspace{-5pt}
\section{Introduction}

Since the introduction of AlexNet~\cite{krizhevsky2012alexnet}, there has been an exponential increase in the number of exceptional convolutional neural networks proposed, resulting in promising outcomes for a variety of visual tasks~\cite{krizhevsky2012alexnet,long2014segmentation,ren2015frcnn,toshev2014humanpose,zhu2016action}. Despite the remarkable results, deploying CNN models on embedded or mobile devices proves challenging as it poses an immense burden on computation and memory storage. To address this issue, a significant amount of research has been dedicated to reducing associated costs, thereby making CNN models more practical for real-world applications. Broadly speaking, this line of research can be categorized into three distinct areas: efficient structure design, network pruning, and network quantization.

Efficient structural design is a challenge in research, with introduction of the separated convolution~\cite{howard2017mobilenets} proposed as an effective technique. This method factorizes the standard convolution into a depthwise and pointwise convolution, reducing computation. Successful examples of its use in efficient networks include MobileNets~\cite{howard2017mobilenets,sandler2018mobilenetv2,howard2019searching} and ShuffleNets~\cite{zhang2018shufflenet, ma2018shufflenetv2}. These networks are widely used on resource-constrained devices and have shown promise in practical applications.
Besides that, various pruning strategies~\cite{han2015learning} have also been proposed to reduce both the computational and storage burdens. However, these methods often incur accuracy degradation, making them less attractive for practical applications. As such, the focus has shifted towards developing efficient structural design methods that can provide high accuracy without compromising on computational efficiency.

Quantization is an incredibly efficient method for deploying models. Its effectiveness has been demonstrated in recent work~\cite{courbariaux2016bnn,rastegari2016xnor,jacob2018quantization,liu2018bireal,hu2022elnet} and has made it increasingly popular in the industry due to its hardware-friendly properties, which can significantly reduce computational and memory costs. However, despite its advantages, there exists a significant accuracy gap between a full-precision model and a quantized counterpart. This gap is especially pronounced in low-bitwidth quantization situations (e.g. 4-bits). Nevertheless, researchers are actively working on closing this gap and making quantization even more effective.

This paper presents a systematic exploration of the effects of each quantization factor on convolutional neural networks, forming a gold guideline for post-training quantization. Additionally, our proposed multi-stage fine-tuning strategy enables quantization to work in conjunction with existing pruning strategies. Exhaustive experiments with ResNet-50 and MobileNetV2, quantized using different bitwidths on ImageNet, showcase the effectiveness of our proposed method.
\section{Quantization scheme}

In this section, we have collected, reviewed and improved upon existing methods for obtaining accurate quantized neural networks. We then conducted experiments to examine the effectiveness of these methods using two representative models, ResNet50~\cite{he2016resnet} and MobileNetV2~\cite{sandler2018mobilenetv2}, on the ImageNet dataset~\cite{russakovsky2015imagenet}. The bitwidth of weights and activations is set to 8, unless stated otherwise in the experiments.

The ImageNet classification task is widely recognized as a challenging benchmark for testing the effectiveness of quantization algorithms. This dataset consists of 1.28M training images and 50K validation images. During the full-precision training, we utilized the random-size cropping strategy~\cite{simonyan2014vgg}, which is then followed by random flipping and mean channel subtraction. Our mini-batch size is set to 256, and we use a momentum SGD optimizer. For ResNet50, the initial learning rate is set to 0.1, and we decay it by a factor of 10 at the 30th, 60th, and 90th epoch, respectively, for a total of 100 epochs. For MobileNetV2, the initial learning rate is also 0.1, and we decay it by a factor of 10 at the 90th, 180th and 270th epoch, for a total of 300 epochs.

\vspace{-5pt}
\subsection{Post-training Quantization} \label{sec:ptq}

\noindent\textbf{Determining the optimal quantization scale.} \label{sec:scale}
One of the main challenges in conducting efficient quantization is determining the optimal quantization scale. Choosing a large scale for a given bitwidth allows for wider representation which can avoid data overflow, but also suffers from more distribution errors. Conversely, selecting a smaller scale can reduce the quantization error in most cases, but it may lead to data overflow due to a narrow data range, resulting in numerical calculation errors. To 
tackle this, NVIDIA's TensorRT~\cite{migacz2017trt} employs a post-training quantization approach that sweep quantization scales by minimizing the Kullback-Leibler (KL) divergence between featuremaps distribution before and after quantization. This approach has shown excellent results in quantizing popular CNN models such as ResNet and MobileNet without the need for additional fine-tuning.

In the context of mathematical statistics, the KL-divergence is a measure of the difference between two probability distributions. Therefore, it can be used to determine the appropriate quantization scale by minimizing the KL-divergence between real values and their quantized counterparts. However, we have found through experimentation that directly using the scale corresponding to the minimal KL-divergence often results in inferior performance. Instead, we have discovered that the optimal scale that yields superior results is typically larger than the scale with the minimal KL-divergence. 
Herein, we propose the introduction of a tolerance coefficient $T$ ($T\geq1$) to improve the performance of quantization in CNNs. Rather than selecting the scale with minimal KL-divergence, we sweep a range of scale candidates and choose the maximum value whose corresponding KL-divergence is less than $T$ times the minimal KL-divergence. 

Table~\ref{tab:tolerance} presents the results of our method. For simplicity, we set a global tolerance parameter, $T$, for all layers in the network. A value of 1.0 for $T$ corresponds to using the conventional KL-divergence algorithm for featuremap quantization, which selects the quantization scale with minimal KL-divergence and serves as a baseline for comparison. As $T$ increases, we observe significant performance improvement for both ResNet50 and MobileNetV2. However, if $T$ is set to a very high value, it degenerates into the naive symmetric \textit{MinMax} quantization, which simply selects the maximum value of a tensor as the quantization threshold. We found that setting $T$ to 1.3 yields a significant improvement over most models.

\begin{table}[htb]
	\renewcommand\arraystretch{1.0}
	\captionsetup{font=small}
	\centering
	\resizebox{0.8\linewidth}{!}{
    	\setlength{\tabcolsep}{2.0mm}{
        	\begin{tabular}{ccccccc}
        		\hline
        		& \multicolumn{2}{c}{\quad  ResNet50 \quad \quad \quad} & \multicolumn{2}{c}{MobileNetV2 \quad} & \\ \cmidrule(lr){2-3} \cmidrule(lr){4-5}
        		$T$ & Top-1 & Top-5 & Top-1 & Top-5 \\ \hline
        		1.0 & 74.88 & 92.45 & 69.74 & 88.98  \\ 
        		1.1 & 75.37 & 92.73 & \textbf{71.07} & \textbf{89.70} \\ 
        		1.2 & 76.05 & 93.02 & 70.82 & 89.63  \\ 
        		1.3 & \textbf{76.20} & \textbf{93.10} & 70.91 & 89.65 \\ 
        		1.4 & 76.17 & 93.07 & 70.80 & 89.56  \\ 
        		1.6 & 76.10 & 93.05 & 70.69 & 89.55  \\ 
        		100.0 & 76.00 & 92.99 & 70.49 & 89.37  \\ \hline
        		FullPrec & 76.39 & 93.18 & 71.93 & 90.30 \\ \hline
        	\end{tabular}
    	}
	}
    \vspace{-5pt}
	\caption{The impact of tolerance coefficient with ResNet50 and MobileNetV2.}\label{tab:tolerance}
	\vspace{-10pt}
\end{table}

\vspace{1pt}
\noindent\textbf{Granularity of quantization.} 
In general, using a single quantization scale for the entire tensor often yields satisfactory results. However, a recent study by Jacob et al.~\cite{jacob2018quantization} has shown that leveraging separate scales for each kernel within the convolutional weights can yield excellent performance, without adding to the computational complexity when deploying the model on hardware. In the literature, weight quantization in convolutional layers is classified into two types: layer-wise and channel-wise quantization. While only layer-wise quantization is allowed for activations, channel-wise quantization can be applied to convolutional weights.

Table~\ref{tab:channel-wise} presents a comparison of layer-wise and channel-wise quantization for 8-bits and 7-bits, respectively. The results show that channel-wise quantization consistently outperforms layer-wise quantization in all cases, which is to be expected as it offers greater flexibility with less distribution error. Additionally, the performance gap becomes more pronounced as the bitwidth decreases. For instance, in 8-bits quantization, MobileNetV2 achieves a top-1 accuracy of 71.45\% with channel-wise quantization, which outperforms the layer-wise variant by 0.54\%. However, the gap widens to 2.48\% in 7-bits quantization. These results demonstrate the superiority of channel-wise quantization for weights in neural networks.

\begin{table}[htb]
	\renewcommand\arraystretch{1.0}
	\captionsetup{font=small}
	\centering
	\resizebox{0.85\linewidth}{!}{
    	\setlength{\tabcolsep}{4.0mm}{
        	\begin{tabular}{ccccc}
        		\hline
        		& \multicolumn{2}{c}{ResNet50}& \multicolumn{2}{c}{MobileNetV2}\\ \cmidrule(lr){2-3} \cmidrule(lr){4-5} 
        		Precision & LW & CW & LW & CW \\ \hline
        		8 & 76.20  & \textbf{76.32}  & 70.91  & \textbf{71.45} \\ 
        		7 & 73.69  & \textbf{75.21}  & 67.69  & \textbf{70.17} \\ \hline
        		FullPrec & 76.39  & 76.39  & 71.93  & 71.93  \\ \hline
        	\end{tabular}
    	}
	}
	\vspace{-5pt}
	\caption{Top-1 accuracy comparison of layer-wise (LW) and channel-wise (CW) quantization for weights in 8-bits and 7-bits.} \label{tab:channel-wise}
	\vspace{-10pt}
\end{table}

\vspace{5pt}
\noindent\textbf{Unsigned quantization:} ReLU activation is commonly employed in modern architectures to combat the issue of gradient vanishing. In ReLU, negative activations are clipped to zero. When quantizing the featuremap after ReLU, unsigned quantization can be applied to improve the representation of positive values by discarding the negative ones. Essentially, this approach is equivalent to adding an additional bit to the quantization of positive values, which effectively doubles the range of representation.

To assess the effectiveness of unsigned quantization, we compared its results with signed quantization. As shown in Table~\ref{tab:pos_quant}, we observed that the type of quantization had a significant impact on efficient networks. With the use of unsigned quantization, MobileNetV2 achieved a top-1 accuracy of 71.94\%, which is equal to the accuracy of full-precision training. On the other hand, the variant with signed quantization only achieved a top-1 accuracy of 71.28\%. Therefore, the inclusion of unsigned quantization for positive activations is crucial to achieve better accuracy at no extra cost.

\begin{table}[htb]
	\renewcommand\arraystretch{1.0}
	\captionsetup{font=small}
	\centering
	\resizebox{0.87\linewidth}{!}{
    	\setlength{\tabcolsep}{1.0mm}{
        	\begin{tabular}{lcccccc}
        		\hline
        		& \multicolumn{2}{c}{Signed} & \multicolumn{2}{c}{Unsigned} & \multicolumn{2}{c}{FullPrec} \\ \cmidrule(lr){2-3} \cmidrule(lr){4-5} \cmidrule(lr){6-7}
        		& Top-1 & Top-5 & Top-1 & Top-5 & Top-1 & Top-5 \\ \hline
        		ResNet50 & 76.19 & 93.05 & \textbf{76.23} & \textbf{93.06} & 76.39 & 93.18\\ 
        		MobileNetV2 & 71.28 & 89.82 & \textbf{71.94} & \textbf{90.18} & 71.93 & 90.30\\ 
        		\hline
        	\end{tabular}
    	}
	} 
	\vspace{-5pt}
	\caption{Comparison of signed and unsigned quantization for activations.}
	\label{tab:pos_quant}
	\vspace{-10pt}
\end{table}

\vspace{2pt}
\noindent\textbf{Quantization Placement} 
Typically, quantization follows convolutional layers, with or without activation layers. In residual-like networks that have shortcut connections via element-wise summation, it is common to quantize both the output activations of the residual branch and the features after summation. However, we have observed that performing quantization at the end of the residual branch can significantly degrade performance. This is because the quantization scale of the two input features needs to be consistent in the addition case. We consider this a hard constraint that can cause significant accuracy degradation. Therefore, we prefer to keep floating-point arithmetic (e.g., half-precision) in element-wise summation and use it in all of our experiments.

\vspace{5pt}
\noindent\textbf{Accumulation in INT16.} \label{sec:accumulation}
To avoid data overflow in the accumulation process of convolution, it is common practice to use the INT32 data type for storing intermediate accumulation results, even though weights and activations are quantized to 8-bits or fewer bitwidth. However, to further reduce the latency and memory footprint, we propose using the INT16 data type for accumulation when the summation of the bitwidth of weights and activations is 14 or less. In our settings, we quantize the weights of convolutions to 6 bits and the activations to 8 bits, which meets this requirement. 

Table~\ref{tab:int16} presents the results. We find that the accuracy is almost retained when replaced with INT16 accumulation in quantized MobileNetV2, which validates the efficacy of using INT16 for accumulation.

\begin{table}[htb]
	\renewcommand\arraystretch{1.0}
	\captionsetup{font=small}
	\centering
	\resizebox{0.68\linewidth}{!}{
    	\setlength{\tabcolsep}{4.0mm}{
        	\begin{tabular}{ccc}
        		\hline
        		& Top-1 & Top-5 \\ \hline
        		Accum\_INT32 & 73.32 & 91.16 \\
        		Accum\_INT16 & 73.26 & 91.11 \\
        		FullPrec & 72.91 & 90.82 \\ \hline
        	\end{tabular}
    	}
	}
	\vspace{-5pt}
	\caption{Comparison of accumulation in INT32 and INT16 of quantized MobileNetV2. The activations and weights are quantized to 8-bits and 6-bits separately.}
	\label{tab:int16}
	\vspace{-10pt}
\end{table}

\vspace{2pt}
\noindent\textbf{Guideline for Post-training quantization.} In summary, to improve accuracy for post-training quantization, we propose the following guidelines:
\vspace{-3pt}
\begin{itemize}
	\item Use an improved KL-divergence algorithm with a tolerance coefficient to determine the scales of activations, and apply \textit{MinMax} quantization to determine the scales of weights. \vspace{-7pt}
	\item Use channel-wise quantization for weights, which is superior to layer-wise quantization, especially for efficient convolutions. \vspace{-7pt}
	\item Employ unsigned quantization on positive activations, such as the output of ReLU. \vspace{-7pt}
	\item Eliminate quantization for the inputs of addition in a network whenever possible, as this can result in significant gains with only a slight increase in computation.
\end{itemize}
\vspace{-3pt}

The results of post-training quantization are shown in Table~\ref{tab:post-training}. For comparison, we used the commercial, closed-source inference library TensorRT and report the average results of 8 experiment runs that used different calibration data. By following the guidelines outlined above, we were able to achieve full-precision accuracy levels in both ResNet50 and MobileNetV2 when quantized to 8 bits. In fact, our approach outperformed TensorRT by a small margin.

\begin{table}[htb]
	\renewcommand\arraystretch{1.0}
	\captionsetup{font=small}
	\centering
	\resizebox{0.86\linewidth}{!}{
    	\setlength{\tabcolsep}{1.5mm}{
        	\begin{tabular}{lcccc}
        		\hline
        		& \multicolumn{2}{c}{TensorRT} & \multicolumn{2}{c}{\textbf{Ours}} \\
        		\cmidrule(lr){2-3} \cmidrule(lr){4-5} 
        		& Top-1 & Top-5 & Top-1 & Top-5 \\ \hline
        		ResNet50 (INT8) & 76.25 & 93.10 & \textbf{76.35} & \textbf{93.13} \\
        		ResNet50 (FullPrec) & 76.39 & 93.18 & 76.39 & 93.18 \\ \hline
        		MobileNetV2 (INT8) & 72.47 & 90.51 & \textbf{72.67} & \textbf{90.64}\\
        		MobileNetV2 (FullPrec) & 72.91 & 90.82 & 72.91 & 90.82\\ \hline
        	\end{tabular}
    	}
	}
	\vspace{-5pt}
	\caption{Accuracy comparison of our method with TensorRT.}
	\vspace{-10pt}
	\label{tab:post-training}
\end{table}

\vspace{-10pt}
\subsection{Quantization-Aware Training}

\vspace{-2pt}
\noindent\textbf{Batch Normalization folding.} Batch Normalization (BN)~\cite{ioffe2015bn} is a crucial module that enhances the stability of training and facilitates optimization of deeper neural networks with minimal weight initialization requirements. During training, a BN operation is supposed to be a linear operation that can be absorbed into the previous convolutional layer. 

To enable quantization-aware training, we need to implement BN folding to determine the optimal scale during forward pass. The gradients are then back-propagated in an unfolding manner, allowing us to separately update the previous convolutional weights as well as the scales and shifts of BN. The parameter folding strategy is formulated as below:
\begin{equation}
\hat{W} = \dfrac{\gamma * W}{\sqrt{\delta^{2}+\epsilon}},\; \hat{B} = \beta - \dfrac{\gamma * \mu}{\sqrt{\delta^{2}+\epsilon}},\; S_{\hat{W}} = \dfrac{\gamma * S_W}{\sqrt{\delta^{2}+\epsilon}}
\end{equation}
Where $W$ as the weights of the previous convolution layer,  $\hat{W}$ and $\hat{B}$ as the combined weights and biases. $\mu$, $\sigma^2$, $\gamma$ and $\beta$ are the mean, variance, scale and shift value of BN, respectively. $S_W$ and $S_{\hat{W}}$ are the quantization scale vector of convolutional weights before and after combination. It's worth noting that $S_{\hat{W}}$ only needs to be executed once after training.

\vspace{5pt}
\noindent\textbf{Quantization-Aware Training.} 
While quantization of the weights and activations in a lower precision, such as 4-bits, the post-training quantization can not maintain the accuracy with a tolerance loss. In this situation, we fine-tune the quantized model to further improve the accuracy by enhancing the fitness and robustness of weights for quantization.

We first quantize a well-trained full-precision model according to the post-training guideline. Then, we fine-tune it with a modified training procedure that adapts the model to quantization. We disable aggressive random-size cropping and use weaker random cropping augmentation. The models are fine-tuned for 20 epochs with a fixed quantization scale for weights, improving training convergence. The initial learning rate is 5e-4, decayed by 5 at the 10th epoch, achieving comparable performance.

Table~\ref{tab:finetune} shows the comparison results on MobileNetV2. It is challenging to achieve quantization with lower precision, but our method achieves the best performance in 4-bits, 5-bits, and 6-bits quantization, surpassing other methods.

\begin{table}[htb]
	\renewcommand\arraystretch{1.0}
	\captionsetup{font=small}
	\centering
	\resizebox{0.85\linewidth}{!}{
    	\setlength{\tabcolsep}{2.0mm}{
        	\begin{tabular}{lcccc}
        		\hline
        		& Weight & Activation & Top-1 & Top-5 \\ \hline
        		PACT ~\cite{choi2018pact,wang2019haq} & 4 & 4 & 62.44 & 84.19 \\
        		DSQ ~\cite{gong2019differentiable} & 4 & 4 & 64.80 & - \\
        		\textbf{Ours} & 4 & 4 & \textbf{66.10} & \textbf{86.97}\\ \hline
        		PACT ~\cite{choi2018pact,wang2019haq} & 5 & 5 & 68.84 & 88.58 \\
        		\textbf{Ours} & 5 & 5 & \textbf{70.06} & \textbf{89.19} \\ \hline
        		PACT ~\cite{choi2018pact,wang2019haq} & 6 & 6 & 71.25 & 90.00 \\
        		\textbf{Ours} & 6 & 6 & \textbf{71.52} & \textbf{89.96} \\ \hline
        		FullPrec & - & - & 71.93 & 90.30 \\ \hline
        	\end{tabular}
    	}
	}
	\vspace{-5pt}
	\caption{Comparison result of quantization-aware training on MobileNetV2 in lower bitwidth.}
	\vspace{-10pt}
	\label{tab:finetune}
\end{table}

\begin{table}[htb]
	\renewcommand\arraystretch{1.0}
	\captionsetup{font=small}
	\centering
	\resizebox{0.87\linewidth}{!}{
    	\setlength{\tabcolsep}{3.0mm}{
        	\begin{tabular}{cccccc}
        		\hline
        		& \multicolumn{2}{c}{$1^{st}$ Finetune ($M_2$)} & \multicolumn{2}{c}{$2^{nd}$ Finetune ($M_4$)} \\ \cmidrule(lr){2-3} \cmidrule(lr){4-5}
        		Sparsity  & Top-1 & Top-5 & Top-1 & Top-5 \\ \hline
        		0.0 & 73.53 & 91.27 & 73.35 & 91.11 \\ 
        		0.1 & 73.50 & 91.34 & 73.34 & 91.21 \\ 
        		0.2 & 73.46 & 91.31 & 73.21 & 91.16 \\ 
        		0.3 & 73.25 & 91.27 & 73.11 & 91.02 \\ \hline
        		FullPrec & 72.91 & 90.82 \\ \hline
        	\end{tabular}
    	}
	}
	\vspace{-5pt}
	\caption{Result of fine-tuning MobileNetV2 with different levels of sparsity in full-precision and the subsequent quantization-aware training in 8-bits.}
	\vspace{-15pt}
	\label{tab:sparsity}
\end{table}

\vspace{5pt}
\noindent\textbf{Integration with pruning.} We now explore the integration of quantization with network pruning strategies that aim to accelerate inference and reduce memory footprint as well. We utilize a simple non-structured pruning approach that zeroes out the ``unimportant" weights with relatively small absolute values, as the design of the sparsity strategy is not our main focus.
To incorporate pruning with quantization, we propose a pipeline that consists of the following steps: \textit{Sparsity} $\rightarrow$ \textit{Full-precision Fine-tuning} $\rightarrow$ \textit{Post-Training Quantization} $\rightarrow$ \textit{Quantization-Aware Training}. Specifically, we first apply the pruning strategy to a well-trained full-precision model $M_1$, which maintains its sparsity throughout the 
entire pipeline.
Following the pruning step, we fine-tune the model with a small learning rate, which allows for minor adjustments to the remaining parameters and generates another sparse full-precision model $M_2$. The next step involves post-training quantization, followed by quantization-aware training, as outlined above. This results in two successive quantized models, namely $M_3$ and $M_4$.

We conducted experiments on the efficient MobilNetV2, and the results are shown in Table~\ref{tab:sparsity}. We first trained a full-precision model with a top-1 accuracy of 72.91\%, then pruned the model with 0\%, 10\%, 20\%, and 30\% sparsity ratios. Interestingly, after the first fine-tuning, we observed that all pruned models outperformed the full-precision baseline, including the model with no pruning (Sparsity=0). This may be attributed to the weakening of data augmentation during fine-tuning, as in quantization-aware training. After quantization-aware training (i.e., the second fine-tuning), we obtained a quantized model with a top-1 accuracy of 73.35\% without sparsity, the best performance for an 8-bits MobileNetV2 to date. Additionally, all quantized models with different prune ratios performed better than the full-precision counterpart. These results demonstrate that the quantization scheme can work well with network pruning strategies.
\section{Conclusion}

In this work, we first systematically refine and improve the effective methods for post-training quantization and establish a gold guideline in practice. Following this guideline, the models in 8-bits  quantization can reach the accuracy of full-precision counterparts without the need for additional training. Additionally, we propose a simple and efficient quantization-aware training strategy that improves accuracy even further in lower precision settings. Lastly, we demonstrate that quantization can work in conjunction with network pruning strategies and propose a multi-stage fine-tuning pipeline to chain them together. These contributions pave the way for more efficient and accurate deep neural network models in resource-constrained environments.
\bibliographystyle{IEEEbib}
\bibliography{egbib}

\appendix

\section{Details for determining activation quantization scale}

\begin{figure}[htb]
	\captionsetup{font=small}
	\begin{center}
		\includegraphics[trim={0.7cm 8cm 1cm 8cm},clip,width=0.4\textwidth]{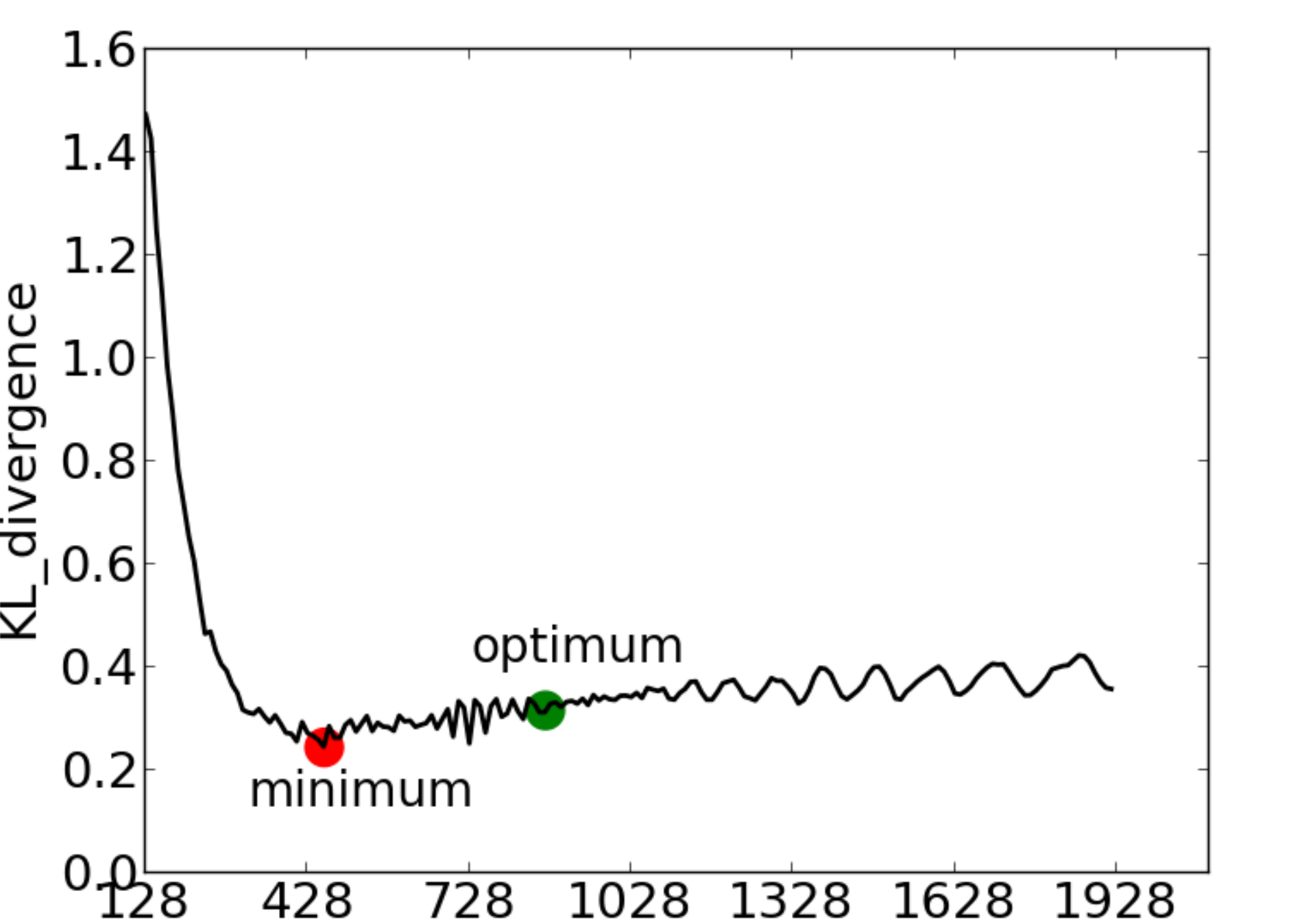}
	\end{center}
	\caption{Distribution of KL-divergence with different quantization scales. The Xß-axis represents the indices of the bins in Algorithm \ref{algo:kl}, where larger quantization scales are represented by higher values. The minimum KL-divergence point is marked by a red dot, while the optimal scale point is marked by a green dot. From this illustration, it is evident that the scale corresponding to optimal performance is always greater than the scale corresponding to minimum KL-divergence.}
	\label{fig:KL_curve}
\end{figure}

Through our experimentation, we have found that the optimal scale for superior results is often greater than the scale corresponding to the minimum KL-divergence as depicted in Fig.~\ref{fig:KL_curve}. Based on this observation, we have enhanced the standard KL divergence algorithm proposed in ~\cite{migacz2017trt}, which was introduced earlier in Sec.~\ref{sec:ptq}. A detailed description of our improved algorithm is provided in Algo.~\ref{algo:kl}.

\begin{algorithm}[!ht]
	\caption{Pseudocode of improved scale determination algorithm for 8-bits layer-wise activation quantization.}
	\label{algo:kl}
	\begin{algorithmic}[1]
		\REQUIRE ~~\\ 
		$ M $: A well-trained full-precision model;\\ 
		$ B $: Number of batches for calibration;\\
		$ T $: Tolerance coefficient;
		\ENSURE ~~\\ 
		N scales for quantized activation layers; \\
		
		\STATE Insert N quantized layers after each activation layer.
		\FOR{each $i \in [1,N]$}
		\FOR{each $j \in [1,B]$}
		\STATE $Activation[i, j] = \textbf{forward}(M, i, j)$ 
		\STATE $DataCollect[j] = \textbf{abs}(Activation[i, j]) - \{0\} $
		\ENDFOR
		\STATE $MaxValue = \textbf{max}\ (DataCollect)$
		\STATE $WidthBins = MaxValue\ /\ 2048$
		\STATE $Bins[2048] = \textbf{histgram}\ (DataCollect, 2048)$
		\FOR{each $ j \in [128, 2048] $}
		\STATE $MergeBins = \textbf{float}(j\;/\;128.0)$
		\STATE $Bins[128] = \textbf{merge}(Bins[2048], MergeBins)$
		\STATE $DBins[2048] = \textbf{map}(Bins[128], MergeBins)$
		\STATE $KL_j = \textbf{cal\_kl\_div}(DBins[2048], Bins[2048])$
		\ENDFOR
		\STATE $ KL_{min} $, $ID_{min} = \textbf{min}\;(KL)\;$, $\textbf{argmin}\;(KL)$
		\STATE $ ID_{opt}  =  ID_{min} $
		\FOR{each $ j \in [128, 2048] $}
		\IF{$KL_j \leq T * KL_{min}$ and $j > ID_{opt}$}
		\STATE $ID_{opt} = j$
		\ENDIF
		\ENDFOR
		\STATE $ scale_{i} = (ID_{opt} + 0.5) * WidthBins / 128$
		\ENDFOR
		\STATE Return [$scale_{1}$, $scale_{2}$, ..., $scale_{N}$]
	\end{algorithmic}
\end{algorithm}

\begin{figure}[!htb]
	\captionsetup{font=small}
	\begin{center}
		\includegraphics[trim={350, 35, 270, 0},clip,width=0.25\textwidth]{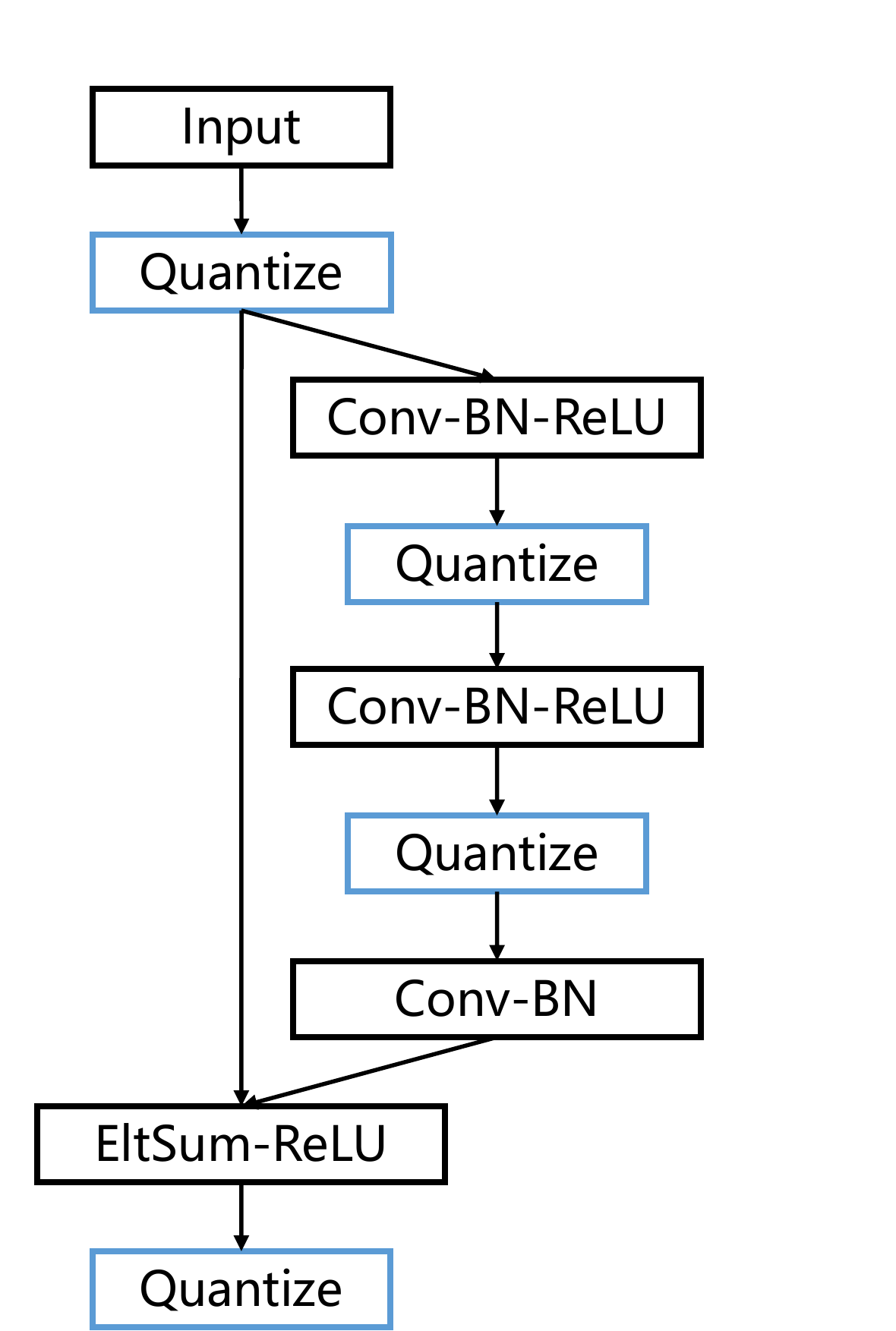}
	\end{center}
	\caption{The schema of quantized bottleneck block in ResNet.}
	\label{fig:residule}
\end{figure}

\begin{figure}[!htb]
\captionsetup{font=small}
    \begin{center}
        \includegraphics[trim={0.2cm 8cm 2cm 2cm},clip, width=0.5\textwidth]{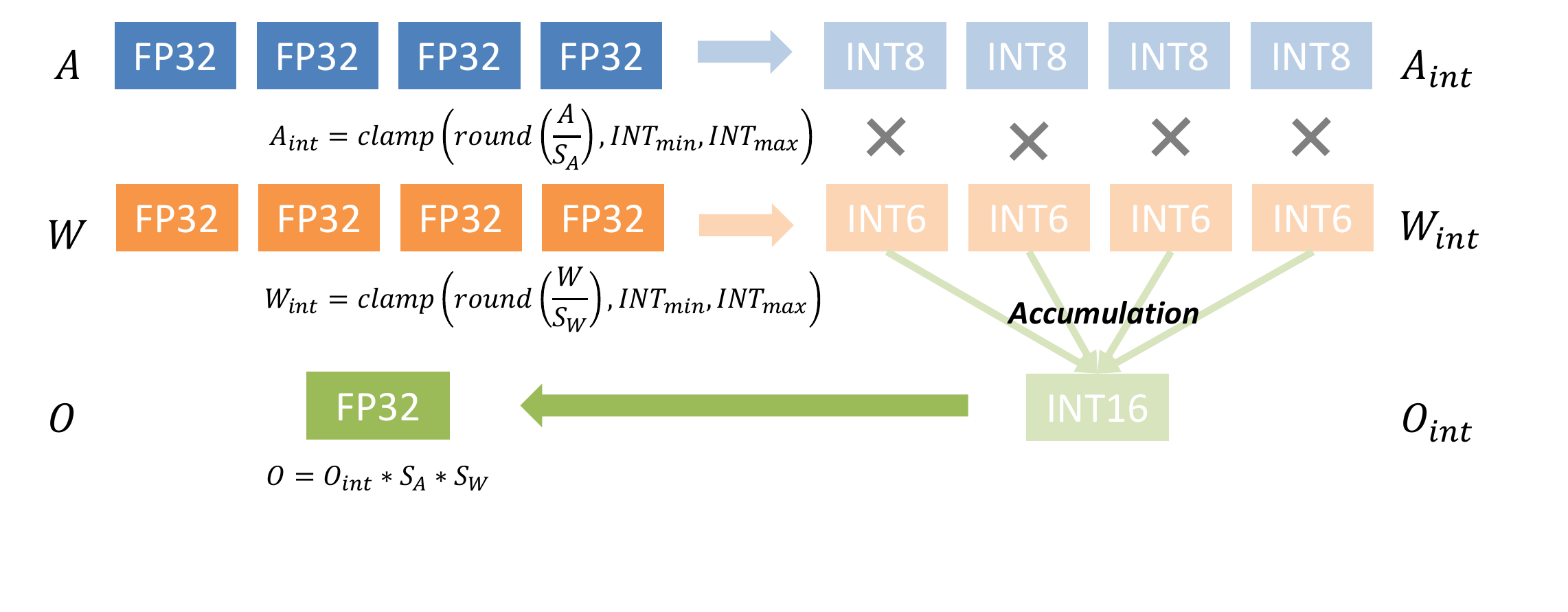}
    \end{center}
    \vspace{-20pt}
    \caption{The schematic of a matrix dot product within a convolution.}\label{fig:accumulation}
\end{figure}

\section{Illustration for Quantization Placement}

As mentioned before, we prefer to utilize floating-point arithmetic (such as half-precision) for element-wise summation. An example of a quantized bottleneck block in ResNet is illustrated in Fig.~\ref{fig:residule}.

\section{Elaboration on Accumulation in INT16}

In later Sec.~\ref{sec:ptq}, we have briefly mentioned the accumulation process. Here, we aim to provide more details about this process. Figure~\ref{fig:accumulation} illustrates the accumulation process of a quantized convolution, where $A$, $W$ and $O$ denote the floating-point input activations, weights and output activations of a convolution layer. A quantized convolution operation is performed wherein $A$ and $W$ are initially quantized using quantization scales $S_A$ and $S_W$ to INT8 and INT6 representations, respectively, denoted by $A_{int}$ and $W_{int}$. Conventionally, INT32 is used to store the accumulation to avoid the possibility of data overflow. However, our experiments have revealed that the more memory-efficient INT16 data type suffices to perform this task. To recover the floating-point result, one can simply multiply the accumulation in INT16 with the product of the quantization scales of the activation and weights utilized for their respective quantization.

\end{document}